\documentclass[11pt]{article}

\usepackage[preprint]{acl}

\usepackage{times}
\usepackage{latexsym}
\usepackage[T1]{fontenc}
\usepackage[utf8]{inputenc}

\usepackage{microtype}
\usepackage{inconsolata}
\usepackage{graphicx}

\usepackage{amsmath,amssymb}
\usepackage{algorithm}
\usepackage{algorithmic}
\usepackage{bm}
\usepackage{multirow}
\usepackage{booktabs}
\usepackage{listings}
\usepackage{xcolor}
\usepackage{tcolorbox}
\tcbuselibrary{breakable}
\usepackage{bbding}
\usepackage{xurl}
\usepackage{enumitem}
\usepackage{colortbl}
\usepackage{CJKutf8}
\usepackage{url}

\lstdefinelanguage{json}{
    basicstyle=\ttfamily\footnotesize,
    stringstyle=\color{teal!70!black},
    keywordstyle=\color{purple!70!black}\bfseries,
    morestring=[b]",
    morekeywords={true,false,null},
    showstringspaces=false,
    breaklines=true,
    breakatwhitespace=true,
    columns=fullflexible,
    keepspaces=true,
    frame=single,
    framerule=0.3pt,
    rulecolor=\color{gray!50},
    backgroundcolor=\color{gray!4},
    xleftmargin=4pt,xrightmargin=4pt,
    aboveskip=4pt,belowskip=4pt,
}

\newenvironment{zh}{\begin{CJK}{UTF8}{gbsn}\itshape}{\end{CJK}}
\newcommand{\benchname}{\textsc{LivingScreen}}

\title{Benchmarking Living-Screen-Native GUI Agents on Short-Video Platforms}

\author{
    Jiashu Yao\textsuperscript{\rm 1},
    Heyan Huang\textsuperscript{\rm 1},
    Daiqing Wu\textsuperscript{\rm 2},
    Wangke Chen\textsuperscript{\rm 1},
    Huaxi Ai\textsuperscript{\rm 1},
    Haoyu Wen\textsuperscript{\rm 1},
    \\
    \bf
    Zeming Liu\textsuperscript{\rm 3},
    Yuhang Guo\textsuperscript{\rm 1}\thanks{Corresponding author.}
    \\
    \textsuperscript{\rm 1}Beijing Institute of Technology \
    \textsuperscript{\rm 2}Tsinghua University \
    \textsuperscript{\rm 3}Beihang University \\
}

\begin{document}
\maketitle
\begin{abstract}
GUI agents today assume a static screen, where the world is frozen between two actions. However, real interfaces such as short-video applications violate this assumption, as their content keeps playing, and a competent user must decide what to watch and for how long. We formalize this task as \textit{Living-Screen-Native} GUI agents and introduce \benchname{}, the first benchmark instantiating it on short-video platforms, with a faithful browser-based environment, a three-tier task suite, and metrics that jointly score accuracy and information efficiency. Evaluating extensive frontier models, we find that none reaches the human cost-accuracy performance, and that their dominant failure mode is over- and under-observation, pointing to observation control as a missing capability axis for future GUI agents. All data and code will be available at \url{https://github.com/BITHLP/LivingScreen}.
\end{abstract}

\section{Introduction}

Multimodal large language models (MLLMs) have rapidly grown into a unified visual understanding~\cite{alayrac2022flamingo, liu2023visual, li2023blip, bai2025qwen3}, where a model now answers questions over still images (VQA)~\cite{antol2015vqa, lu2024mathvista, yue2024mmmu, liu2024mmbench, fu2026mme}, reasons about temporal content in videos (Video QA)~\citep{fu2025video, li2026crossvid, wang2026livibench}, and most recently, drives software interfaces as a general-purpose GUI agent. GUI agents perceive the screen, plan, and issue clicks to accomplish user goals on websites, mobile apps, and desktop OSes~\citep{hong2024cogagent, wang2024mobile, qin2025ui}, turning an MLLM from a passive answer-machine into a deployable assistant.

\begin{figure}[htbp]
    \centering
    \includegraphics[width=1.0\linewidth]{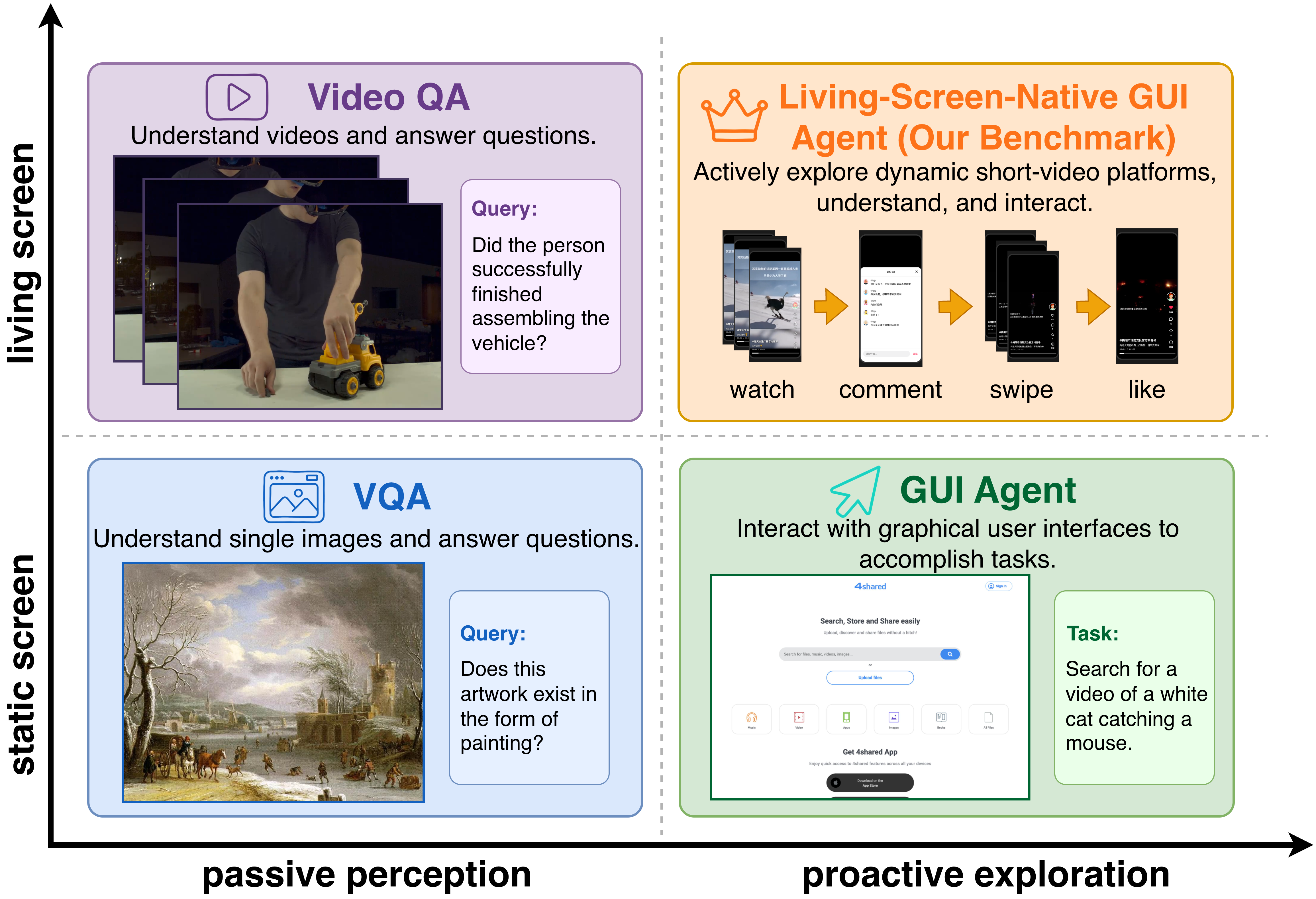}
    \caption{Positioning living-screen-native GUI agents. The vertical axis indicates whether the environment evolves autonomously, and the horizontal axis indicates whether the agent acts on it. Living-screen-native agents (top-right) operate in the only quadrant that combines both, a regime that short-video platforms naturally inhabit but existing benchmarks overlook.}
    \label{fig:intro}
\end{figure}

However, an important piece of this frontier is missing. There is, at present, no benchmark challenges a GUI agent to handle \emph{dynamic video content in its native, on-screen form}, which human users routinely do. Concretely, existing benchmarks either strip the GUI away and directly feed videos files to the model as in Video QA, or strip the video away as in GUI agent benchmarks.

The upper-right quadrant of Figure \ref{fig:intro}, which represents a living screen observed proactively, becomes a gap and our work's target. Operating natively on such a living screen gives the agent unique and important capabilities. First, it allows an agent to proactively choose its own observation granularity, ranging from a single screenshot to a multi-second clip at a chosen frame rate over different content. Second, it operates without needing access to the underlying video file. Third, it consumes video the way the platform actually presents it, co-rendered with comments, captions and bios.

We propose to call agents in this missing quadrant \emph{living-screen-native}. A living-screen-native GUI agent is a GUI agent that operates on a screen evolving in continuous time, and that actively determines what visual slice of that evolving screen to observe. We instantiate the setting on short-video platforms, where the agent's screen is a temporally-evolving canvas dominated by playing video content, co-rendered with static UI affordances (like, comment, share, follow buttons, progress bars), user-generated overlays (comments, captions), and author meta (bio, hashtags). Moreover, our agent issues observations as first-class actions, at each decision point it decides not only where to click but also what to look at, from taking a single screenshot to recording a $\Delta t$ second clip of a certain video at a chosen frame rate. This makes information acquisition an endogenous and cost-bearing decision rather than an exogenous data feed.

To make this setting concrete and measurable, we introduce \benchname{}, the first benchmark for living-screen-native GUI agents on
short-video platforms. \benchname{} consists of three components.
(i) A high-fidelity browser-based environment that faithfully reproduces the affordances of a modern short-video app, and exposes them to agents through a Playwright-based action API.
(ii) A three-tier task suite progressing from L1 atomic GUI operations, through L2 cross-source understanding, to L3 closed-loop applications.
(iii) A set of evaluation metrics that score both task accuracy and information efficiency.

We comprehensively evaluate frontier MLLMs as agents on \benchname{} and find that the setting is highly challenging, as even the strongest model trails human performance by a wide margin on the cost-accuracy tradeoff. A closer analysis traces the dominant failure mode to a previously under-studied phenomenon we call \textit{over- and under-observation}, a visual-channel analogue of over- and under-thinking, in which models systematically watch either far more or far less than the task requires, suggesting that observation control is a genuine capability gap of current MLLMs.

% ---- Para 6: contributions ----
Our contributions are summarized as follows.
\begin{itemize}\itemsep1pt
    \item We identify and formalize \emph{Living-Screen-Native} GUI agents, a setting in which the screen evolves in continuous time and the agent itself decides what visual slice of that screen to observe.
    \item We release \benchname{}, the first benchmark instantiating this setting on short-video platforms, comprising a faithful browser-based environment, a three-tier task suite, and metrics that jointly score accuracy and information efficiency.
    \item We benchmark frontier MLLMs and uncover \textit{over- and under-observation} as the dominant failure mode of current living-screen-native agents, which we argue should be treated as a novel axis for future GUI-agent research.
\end{itemize}

\section{Related Work}

\subsection{Video Understanding Benchmarks}
Benchmarks for video understanding have grown rapidly in scale and difficulty, but they universally adopt a passive consumption protocol, where fixed clips and a question are handed to the model~\cite{fu2025video, wu2024longvideobench, zhou2025mlvu}. A more recent line push the multi-video understanding frontiers for robustness~\cite{wang2025muirbench} cross-video reasoning \cite{zhu2025cvbench, li2026crossvid}. LiViBench~\cite{wang2026livibench} further couples livestream videos with their comments to evaluate interactive understanding. All of these benchmarks present the videos and side information as a flat context, where the model neither chooses what to look at nor how long to look. Our setting replaces such input with a continuously feed and a long-horizon goal, so that information selection itself becomes part of the task.

\subsection{GUI Agents and Their Benchmarks}
GUI agents have evolved from external-component-assisted architectures~\cite{wang2024mobile} to end-to-end vision-grounded ones~\cite{hong2024cogagent, wu2025atlas, qin2025ui, ye2025mobile, xu2026mobile}. The evaluation ecosystem is equally rich, covering web~\cite{zhou2024webarena, koh2024visualwebarena}, mobile~\cite{toyama2021androidenv, rawles2025androidworld}, and desktop control~\cite{xie2024osworld}. Despite their breadth, these environments treat the screen as a discrete sequence of screenshots, where the world is frozen between two agent actions. \benchname{} keeps the screen alive, as short-video content auto-plays in real time, and elevates ``how long and at what granularity to watch'' to a first-class decision, a dimension that conventional GUI benchmarks do not measure.

\subsection{Agentic Video Understanding}
Closest in spirit to our work is a recent line that recasts video QA as an agentic decision problem, where a controller retrieves and zooms into video evidence. Video-Browser~\cite{liang2025video} equips an agent with YouTube search, preview, and detailed-inspection APIs to answer questions. Others \cite{wang2025videotree, yuan2025videodeepresearch, zhang2026deep} let an LLM actively segment, retrieve, or hierarchically prune long videos rather than ingesting them in one pass. These works share our core motivation, that ``which frames to look at'' should be decided by a policy, not by a fixed sampler. However, their action spaces are human-designed previledged tool calls (\texttt{search}, \texttt{zoom}, \texttt{retrieve}) operating over an offline corpus, where the agent never touches a real interface, and the underlying video is still a static asset. \benchname{} requires every act of information gathering to go through pixel-level GUI interaction on a auto-playing feed by swiping the feed, scrubbing the progress bar, and expanding comments. The agent therefore inherits both the flexibility of agentic video exploration and the grounding burden of a GUI agent, and is evaluated under a continuous-time state rather than a discrete retrieve-then-answer loop.

\section{Benchmark}

\begin{figure*}[t]
  \includegraphics[width=1.0\linewidth]{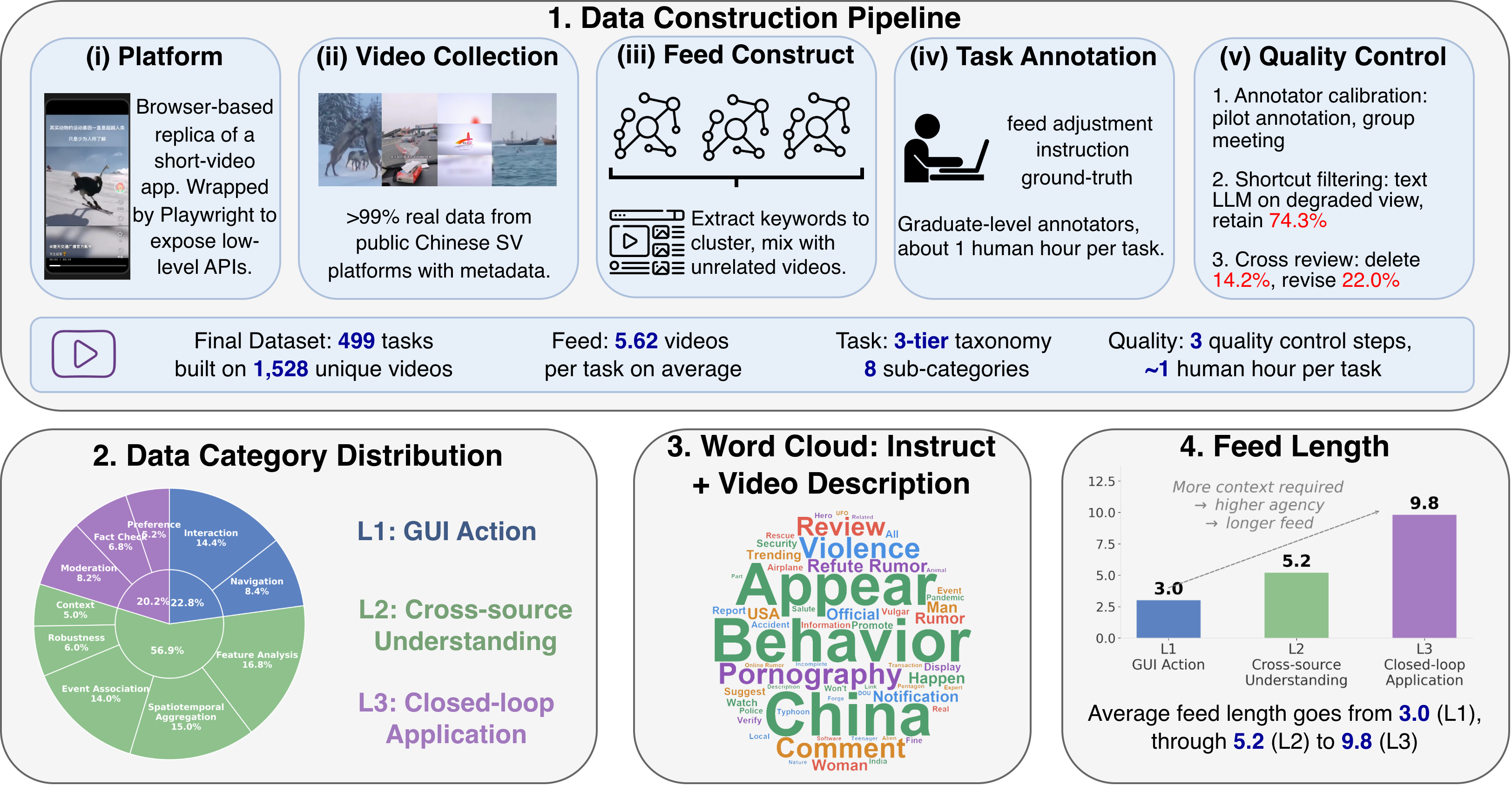}
  \caption{Illustration of \benchname{}.}
  \label{fig:data}
\end{figure*}

\subsection{Task Formulation}

\paragraph{Background}
A standard GUI agent is modeled as a partially-observed Markov decision process (POMDP)
$\mathcal{M}=(\mathcal{S},\mathcal{A}, \mathcal{O}, \mathbb{T}, \mathbb{O}, R, N)$,
where the state evolves only when the agent acts through
$s_{t+1} \sim \mathbb{T}(\cdot\mid s_t,a_t)$, and at every step the
agent receives a single screenshot
$o_t \sim \mathbb{O}(\cdot\mid s_t)$. We argue that this discrete-time,
agent-driven formulation cannot describe video-bearing screens, and
introduce two principled modifications.

\paragraph{($\Delta_1$) Continuous evolving state}

On a video-bearing screen the playing video advances even when the
agent does nothing, so the underlying state cannot be indexed by a
discrete agent-step counter. We therefore lift the state to a
continuous-time stochastic process and decompose its dynamics into two superimposed mechanisms. First, between
two consecutive decision instants $t_k$ and $t_{k+1}$, the state
follows an autonomous flow $\Phi$ that captures
agent-independent evolution (e.g., the playhead advances, content auto-loads)
\begin{equation}
\label{equ:auto-flow}
    s(t) \sim \Phi_{t-t_k}\!\bigl(\cdot \mid s(t_k^{+})\bigr), \quad t \in (t_k,\, t_{k+1}).
\end{equation}
Second, at a decision instant $t_{k+1}$, the agent issues an action $a_k$,
which produces a discrete jump
\begin{equation}
    s(t_{k}^{+}) \sim \mathbb{T} \bigl(\cdot \mid s(t_{k}^{-}), a_k\bigr).
\end{equation}
Here $s(t^{-})$ and $s(t^{+})$ denote the left and right limits. Equation \ref{equ:auto-flow} is what makes the screen living, and it is absent in standard GUI POMDPs.

\paragraph{($\Delta_2$) Agent-initiated observation}

In video-native GUI scenarios, agents proactively control the scope and granularity of its observation. We replace the classical pointwise observation $o_t \sim \mathbb{O}(\cdot\mid s_t)$
with a observation parameterized by the agents' action
\begin{equation}
\label{equ:proactive}
    o_k \sim \mathbb{O}\Bigl(\cdot \mid \{\,s(t)\,:\,t\in W(a_k)\}\Bigr),
\end{equation}
where $W(a_k)\subseteq\mathbb{R}_{\geq 0}$ is a time window
determined by $a_k$, and $\mathbb{O}$ extracts the visual content
that would actually be rendered on the screen during $W_k$. In Equation \ref{equ:proactive}, it is
the agent who decides watch a single screenshot or a certain slice of a video.

Our formulation, the \textit{Living-Screen-Native POMDP}, differs from a standard POMDP in two places: the state evolves on a continuous time axis ($\Delta_1$, Equ \ref{equ:auto-flow}), and observations are agent-initiated queries over states rather than pointwise samples ($\Delta_2$, Equ \ref{equ:proactive}). The two changes collectively explain why benchmarks built on either standard video QA or standard GUI agent formulations cannot evaluate the capabilities studied in this work.

\subsection{Benchmark Construction}

\paragraph{Platform}
We build the environment as a browser-based replica of a modern short-video application, supporting feed scrolling, liking, collecting, comment threading, and reporting on a fixed-resolution canvas matching a typical mobile viewport. To expose this canvas to agents, we wrap the page with Playwright~\cite{playwright} and provide a low-level action API including pixel-coordinate clicks, swipes, keyboard input, etc. The agent observes the environment exclusively through rendered screenshots or screen recordings, never through privileged DOM access or video files.

\paragraph{Video collection}
The vast majority of the videos in \benchname{} are real short videos collected from publicly accessible Chinese short-video datasets~\cite{livebot, fakesv}, together with associated metadata including title, hashtags, like/comment/save counts, and comment threads attached. To cover long-tail content like rare safety risks, we additionally include a small number of synthesized videos~\cite{liu2025videosafetybench}, which account for less than $1\%$ of the total corpus.

\paragraph{Feed construction}
To mimic the mixture of topical consistency and diversity produced by a real recommender, we use the keyword sets from each video and cluster the corpus. For each task, we uniformly choose a ratio $r\%$ between $50\%-200\%$, draw an anchor cluster and combine it with $r\%$ as many randomly-sampled unrelated videos, then shuffle the combined pool to form a feed. This base feed is the base on which task annotators (below) refine the ordering, delete specific videos, or swap in distractors so that the resulting feed exactly supports the intended question.

\paragraph{Task annotation}
All tasks in \benchname{} are authored by graduate-level annotators with an extensive background in multimodal researches. Each annotator is shown a candidate feed and a task taxonomy to choose from (detailed below). They then write a natural-language query together with the ground-truth answer. Answers take one of two
forms: a labelled option for multiple-choice questions, or a declarative state of the environment (e.g., video $i$ is liked and $j$ is reported) for the platform back-end to conduct automatic evaluation. Where the base feed does not yet support the intended query, annotators are permitted to edit it.

\paragraph{Quality control}
We apply a three-stage quality-control pipeline. (i) \textit{Annotator calibration.} Before authoring at scale, every annotator completes a small pilot batch, which is jointly reviewed in a calibration meeting to resolve ambiguity before formal annotation begin. (ii) \textit{Shortcut filtering.} Once a task is written, we run DeepSeek-V4~\cite{deepseekai2026deepseekv4} on a degraded view of the feed that exposes only non-video metadata (title, hashtags, author bio), and discard any task the model can already solve, retaining $74.3\%$ of authored tasks. (iii) \textit{Cross-annotator review.} The tasks left are re-shown to a different annotator who checks and attempts to answer them. Items flagged as ambiguous, poorly worded, or whose answer the reviewer disagrees with are escalated to group decision. After this round we delete a further $14.2\%$ and revise the wording $22.0\%$ of tasks, yielding the final \benchname{} task set.

\subsection{Task Description}

The final \benchname{} consists of $499$ tasks built on top of $1,528$ unique videos, with an average feed length of $5.62$ videos per task. As is shown in Figure~\ref{fig:data}, tasks are organized into a three-tier taxonomy, \textsc{L1} atomic agent action, \textsc{L2} cross-source understanding, and \textsc{L3} closed-loop application.

\paragraph{L1: GUI action}
\textsc{L1} probes whether the agent can correctly execute the elementary GUI primitives that higher-tier tasks build upon. It contains two sub-categories:
\textit{interaction}, which asks the agent to like, collect, comment, or report designated videos, and
\textit{navigation}, which asks the agent to swipe through the feed or to seek along the progress bar to a target timestamp.

\paragraph{L2: cross-source understanding}
\textsc{L2} measures whether the agent can integrate evidence across videos, comments, and other cues within a feed. It comprises five sub-categories, including
\textit{contextual association} (linking videos and comments),
\textit{event association} (comparing storylines across clips),
\textit{feature analysis} (comparing fine-grained attributes across clips),
\textit{spatiotemporal aggregation} (counting or summing quantities across clips), and
\textit{robust evaluation} (a controlled subset of unanswerable questions to probe trustworthiness).

\paragraph{L3: closed-loop application.}
\textsc{L3} closes the loop between perception, reasoning, and action on short-video platform, where the agent must browse, decide, and operate on the feed to produce an environment-graded outcome. We instantiate three applications, including
\textit{fact-checking} where the agent inspects multiple clips and to judge a contested event,
\textit{content moderation} where the agent screens a feed for a specified policy violation and reports correspondingly, and
\textit{preference simulation} where the agent is asked to behave like a user with a stated interest profile.

\subsection{Evaluation metrics}

\paragraph{Accuracy metrics}
We report a task-level Success Rate (\textit{SR}) that uniformly covers all three tiers. For \textsc{L1} and \textsc{L3}, a task is graded by the environment back-end, where the post-episode platform state is compared against the declarative ground-truth specification (e.g.,\ video $i$ liked, video $j$ reported), and \textit{SR}$=1$ only if the state matches exactly. For \textsc{L2} multiple-choice tasks, \textit{SR} is the standard option-match accuracy on the model's \texttt{answer}.

\paragraph{Efficiency metrics}
We additionally report two cost-side metrics that jointly capture how efficiently the agent gathers information. Number of Steps (\textit{NS}) is the average number of tool calls issued per episode and measures \emph{operational} cost. Watch Ratio (\textit{WR}) is the average fraction of total feed runtime that the agent actually records through \textsc{watch} measuring \emph{observational} cost. A proactively efficient agent should achieve a high \textit{SR} while keeping both \textit{NS} and \textit{WR} small.

\begin{table*}[htb!]
    \centering
    \setlength{\tabcolsep}{1.75mm}
    \begin{tabular}{lccc|ccccccccc}
        \toprule
        \multirow{2}{*}{\textbf{Models}} & \multicolumn{3}{c|}{\textbf{Average}} & \multicolumn{3}{c}{\textbf{Action}} & \multicolumn{3}{c}{\textbf{Understanding}} & \multicolumn{3}{c}{\textbf{Application}} \\
        \cmidrule(lr){2 - 4} \cmidrule(lr){5 - 7} \cmidrule(lr){8 - 10} \cmidrule(lr){11 - 13} & \textit{\textbf{SR$\uparrow$}} & \textit{\textbf{NS$\downarrow$}} & \textit{\textbf{WR$\downarrow$}} & \textit{\textbf{SR$\uparrow$}} & \textit{\textbf{NS$\downarrow$}} & \textit{\textbf{WR$\downarrow$}} & \textit{\textbf{SR$\uparrow$}} & \textit{\textbf{NS$\downarrow$}} & \textit{\textbf{WR$\downarrow$}} & \textit{\textbf{SR$\uparrow$}} & \textit{\textbf{NS$\downarrow$}} & \textit{\textbf{WR$\downarrow$}} \\
        \midrule

        Random&
        -&  -&  -&
        -&  -&  -&  25.1&  -&	-&	-&	-&	-
        \\
        Human&
        94.0&   -&   9.7&
        100.0&  -&  3.1&    88.0&   -&  8.4&    94.1&   -&  17.5
        \\
        \midrule
        
        \rowcolor{gray!20} \multicolumn{13}{c}{\it Video-input MLLMs} \\
        Gemini-3.5&
        \textbf{69.3}&   8.0&    11.9&
        \textbf{90.4}&	2.4&	2.4&	60.2&	6.1&	19.3&	\textbf{57.4}&	15.6&	14.1
        \\
        Gemini-3.1&
        66.2&   7.6&    11.6&
        87.7&	\textbf{2.3}&	\textbf{2.0}&	64.2&	5.8&	16.7&	46.8&	14.7&	16.2
        \\
        Seed-2.0&
        64.8&  8.1&  15.5&
        88.6&	2.4&	2.1&	\textbf{68.3}&	5.4&	17.5&	37.6&	16.5&	27.0
        \\
        Seed-1.8&
        65.6&  9.5&  25.1&
        86.8&	3.0&	2.4&	64.4&	6.0&	19.8&	45.5&	19.4&	53.1
        \\
        Seed-1.6&
        43.8&  7.8&  \textbf{5.6}&
        74.6&	4.3&	3.4&	45.8&	\textbf{4.5}&	\textbf{7.0}&	10.9&	14.5&	\textbf{6.5}
        \\
        Qwen-3.6&
        44.9&  10.5&  10.9&
        72.8&	4.4&	3.6&	39.1&	6.1&	14.7&	22.8&	21.0&	14.5
        \\
        Qwen-3.5&
        53.0&  10.2&  14.7&
        76.3&	4.1&	3.7&	43.0&	7.5&	22.6&	39.6&	18.9&	17.9
        \\
        GLM-5V-Turbo&
        42.7&  10.6&  24.1&
        68.4&	3.4&	3.0&	46.8&	7.7&	27.9&	12.9&	20.6&	41.4
        \\
        Kimi-K2.5&
        36.8&  11.7&  20.2&
        43.9&	5.2&	4.4&	55.6&	9.0&	25.6&	10.9&	20.8&	30.6
        \\
        \midrule

        \rowcolor{gray!20} \multicolumn{13}{c}{\it Multi-image-input MLLMs} \\
        Claude-Opus&
        45.6&   9.7&    8.5&
        64.0&	4.8&	4.2&	45.1&	5.8&	10.0&	27.7&  18.5&   11.2
        \\
        Claude-Sonnet&
        38.7&   \textbf{6.0}&    8.3&
        67.5&	3.5&	3.0&	37.2&	5.7&	14.1&	11.3&	\textbf{8.7}&	7.7
        \\
        GPT-5.5&
        29.8&   10.1&   14.4&
        50.9&	3.0&	2.4&	38.4&	6.1&	18.6&	0.0&	21.2&	22.3
        \\
        GPT-5.4&
        25.2&  10.2&  15.2&
        37.7&	2.3&	2.1&	38.0&	5.9&	21.2&	0.0&	22.3&	22.2
        \\

        \bottomrule
    \end{tabular}
    \caption{Main results. The best results among all models are shown in bold.}
    \label{tab:main-table}
\end{table*}

\section{Experiments}

\subsection{Settings}

\paragraph{Models}
We benchmark a representative set of frontier multimodal models that cover the major model families, including native video models including Gemini-3.5-Flash, Gemini-3.1-Pro, Doubao-Seed-2.0-Pro, Doubao-Seed-1.8, Doubao-Seed-1.6, Qwen-3.6-Plus, Qwen-3.5-Plus, GLM-5V-Turbo, Kimi-K2.5, and image models including Claude-Opus-4.6, Claude-Sonnet-4.6, GPT-5.5, GPT-5.4. All models are queried with temperature $0.6$, high reasoning effort, and are prompted in an identical think-then-tool-call format.

\paragraph{Agent implementation}
We wrap every model with a single-agent architecture so that the only varying component is the underlying MLLM. At each decision step the agent receives the observation (a screenshot or a recording clip), performs reasoning, and emits a tool call from the action space. Beyond the standard GUI primitives (\texttt{click}, \texttt{right click}, \texttt{swipe}, \texttt{type}, \texttt{press}), we expose two living-screen-native primitives that directly instantiate our formalization. \texttt{Watch} records the on-screen rendering for a time window at a frame rate realizing the agent-initiated observation (Equation \ref{equ:proactive}), and \texttt{wait} lets the autonomous flow advances (Equation \ref{equ:auto-flow}) allowing the agent to skip uninformative segments. An auxiliary \texttt{mark} primitive lets the agent to calibrate the coordination grounding, and \texttt{answer} emits the final response. To keep the comparison tractable, we cap each episode at $30$ steps, retain all the history while only keep the most recent observation image or video in context. The ablation study of the agent design in conducted (as shown in the following section), validating the bottlenect of agent performance is the capability of underlying MLLMs.

\subsection{Main Results}

The main results are shown in Table \ref{tab:main-table}.

\paragraph{Overall benchmark challenge}

\benchname{} presents a significant challenge for current models, as even advanced models struggle to achieve high \textit{SR}. The living-screen-native tasks especially raise challenges to models lacking native video input and precise GUI grounding.
A fine-grained analysis of the results reveals that GUI action execution and cross-source understanding emerge as two independent dimensions. For example, Qwen-3.6-Plus achieves high precision in GUI action while performs relatively weak in understanding tasks, while Kimi-K2.5 exhibits the opposite. Ultimately, achieving success in closed-source applications demands a holistic mastery of both aspects.

\paragraph{Divergent efficiency paradigms} 
Furthermore, different models exhibit highly divergent operational paradigms handling living-screen-native scenarios, as reflected by the substantial variance in efficiency metrics \textit{NS} and \textit{WR}. 
Specifically, certain models (e.g., Kimi-K2.5) favor an exploration-heavy strategy characterized by a high \textit{NS}, frequently executing multiple steps to retrieve fragmented information across various sources. 
Conversely, other models (e.g., Doubao-Seed-1.8) favor an observation-heavy pattern with a high \textit{WR}, dedicating more time to processing continuous video frames to secure a comprehensive context. 
A few conservative models (e.g., Doubao-Seed-1.6) maintain low scores in both metrics, revealing a behavioral tendency to rely on minimal information to complete tasks.
We will further discuss the behavioral divergence and reveal over- and under- observation problems below.

\subsection{Ablations on Agent Design}

\begin{table}[htb!]
\setlength{\tabcolsep}{2mm}
    \centering
    \begin{tabular}{cccc}
        \toprule
        
        \textbf{Max Steps} & \textbf{Max Retain} & \textbf{Reasoning} & \textbf{\textit{SR}} \\
        \midrule

        \rowcolor{gray!20} \multicolumn{4}{c}{\it Ours implementation} \\
        $30$ & $1$ & Yes & 65.6 \\
        \midrule

        \rowcolor{gray!20} \multicolumn{4}{c}{\it Ablations of max steps} \\
        $40$ & $5$ & Yes & 65.6 \\
        $20$ & $5$ & Yes & 58.2 \\
        \midrule

        \rowcolor{gray!20} \multicolumn{4}{c}{\it Ablations of max retaining feedbacks} \\
        $30$ & $2$ & Yes & 61.1 \\
        $30$ & $3$ & Yes & 64.9 \\

         \rowcolor{gray!20} \multicolumn{4}{c}{\it Ablations of reasoning} \\
        $30$ & $1$ & No & 58.8 \\
        
        \bottomrule
    \end{tabular}
    \caption{Ablation on agent designs.}
    \label{tab:ablations}
\end{table}

To verify that the agent design is not the bottleneck of the reported scores, we ablate its three most consequential design choices on {Doubao-Seed-1.8} along the dimensions of maximum episode length, observation retention, and explicit reasoning. Results are summarized in Table~\ref{tab:ablations}. Across all variants the \textit{SR} fluctuates around our default configuration, indicating that our agent has saturated what the underlying model can deliver.

\section{Analysis: Over- and Under-Observation}

A central design choice in \benchname{} is to expose \texttt{watch} as first-class actions, so that how long and how attentively to look at the screen becomes part of the agent's decision rather than a hyperparameter of the data loader. This section inquiry whether current models actually exercise it well. By analogy with the over- and under-thinking phenomena observed in reasoning, we identify and characterize \textit{over- and under-observation} as a core failure mode of current living-screen-native agents, where models systematically watch either too much or too little.

\subsection{Observation Behavior Analysis}

The main results in Table \ref{tab:main-table} already reveal that humans achieve substantially higher \textit{SR} than any evaluated model while spending a lower \textit{WR}. Here we take a closer look at the gap and analyze how human and model observation behavior actually differ at the per-video level.

\begin{figure}[htbp]
    \centering
    \includegraphics[width=1.0\linewidth]{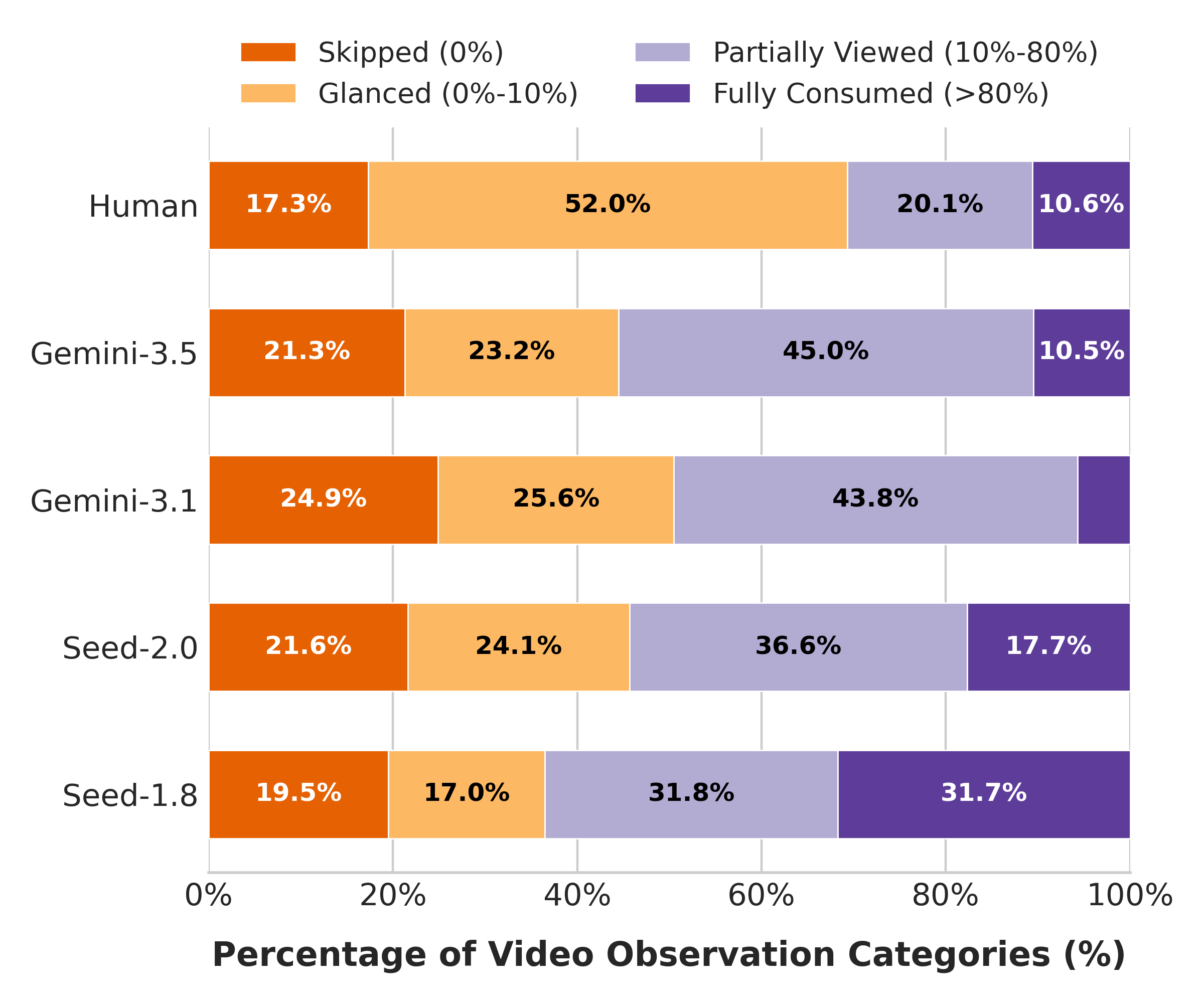}
    \caption{Distribution of per-video observation behaviour across four watch-ratio categories, averaged over all \textsc{L2} and \textsc{L3} tasks.}
    \label{fig:obs-behavior}
\end{figure}

\paragraph{Setup}
We classify the observation made on every (agent, video) pair into four mutually exclusive categories based on its watch ratio $r$. (i) Skipped ($r{=}0\%$) if the video was passed without any attendence, (ii) glanced ($0\%{<}r{\leq}10\%$) if the agent only take a brief look, (iii) Partially Viewed ($10\%{<}r{\leq}80\%$) if there is a substantial but non-exhaustive consumption, and (iv) fully consumed ($r{>}80\%$) if there is a near-complete consumption. The analysis is run over all \textsc{L2} and \textsc{L3} tasks, since \textsc{L1} tasks do not require substantial video content understanding and uniformly lead to a near-zero \textit{WR}. For clarity, we plot the four strongest agents in Figure~\ref{fig:obs-behavior} (Gemini-3.5, Gemini-3.1, Seed-2.0, Seed-1.8) against human annotators on the same tasks.

\paragraph{Comparison among models}
Across the four evaluated agents the overall shape of the distribution is broadly similar, where roughly a quarter of videos are skipped, another quarter are glanced, and the bulk of probability mass sits in partially viewed. The most significant model-to-model difference is the fully consumed part. The two Seed agents, Seed-1.8 in particular, tend to watch videos through to the end ($17.7\%$ and $31.7\%$), while the Gemini agents rarely do ($10.5\%$ and $5.7\%$). This long tail is precisely what inflates \textit{WR} for the Seed models without translating into a proportional \textit{SR} gain in Table~\ref{tab:main-table}.

\paragraph{Comparison between humans and models}
The human row stands apart from all four agents in a single way that humans place over half of their observations ($52.0\%$) in the glanced bucket, whereas no model exceeds $26\%$ there. In other words, human annotators routinely spend a small amount of attention to decide whether a video is worth pursuing, and only then commit to a longer watch. This pattern is consistent with information foraging theory \cite{pirolli1999information}, in which a rational forager interleaves cheap ``patch scouting'' with selective deep consumption to maximize information gain per unit cost. Models, by contrast, appear to lack this cheap scouting step: when an item is engaged at all, it is engaged at length, which explains he lower glanced mass. The same imbalance is also visible on the skipped side. Every model skips more videos than humans do, suggesting that without a reliable glance-based filter, agents fall back to a coarser strategy that simultaneously over- and under-spends attention. We zoom in on these two failure modes in the following.

\subsection{Over- and Under-Observation}

The observation-behavior shown in the previous subsection indicates two complementary failure modes, which are agents looking at far less than the task actually requires, and agents looking at far more than necessary. By analogy with under- and over-thinking in LLM reasoning, we name these two problems \textit{under-observation} and \textit{over-observation}. Note that the two are not mutually exclusive on a single trajectory, as an agent can over-observe some videos in a feed while under-observing others. Here we investigate and quantify the under- and over-observation problems of current MLLMs in living-screen-native tasks.

\paragraph{Setup}
We sample $100$ trajectories per agent, $50$ from \textsc{L2} and $50$ from \textsc{L3}, across Gemini-3.5, Gemini-3.1, Seed-2.0, and Seed-1.8 models. An annotator flag for every video appearing in a sampled trajectory, whether the agent's observation on that video is under-observed (insufficient watching of a video that materially supports the answer) and / or over-observed (substantial watching of a video that is irrelevant to the task or already understood). The per-agent under-observation and over-observation rates are listed in Table~\ref{tab:over-under}.

\paragraph{Both failure modes are pervasive}
Table~\ref{tab:over-under} reports the resulting rates. Under-observation is the more widespread of the two, affecting between $39\%$ and $61\%$ of trajectories across the four agents, and even the strongest agents commit to an answer before information has been examined closely enough on roughly half of the tasks. Over-observation is less frequent but not negligible, ranging from $25\%$ to $43\%$. Crucially, the two are not exclusive, as $9$–$24\%$ of trajectories exhibit both failure modes on the same task, with the agent spending substantial attention on irrelevant videos while skimming the truly informative one. We take this prevalent, two-sided, and frequently co-occurring miscalibration as the core observational failure mode that \benchname{} surfaces in current living-screen-native agents.

\begin{table}[htbp]
    \centering
    \begin{tabular}{lccc}
        \toprule
        \textbf{Agent} & \textit{\textbf{Under$\downarrow$}} & \textit{\textbf{Over$\downarrow$}} & \textit{\textbf{Both$\downarrow$}} \\
        \midrule
        {Gemini-3.5}          & 51 & 29 & 13 \\
        {Gemini-3.1}          & 39 & 25 & 9 \\
        {Seed-2.0}            & 61 & 27 & 20 \\
        {Seed-1.8}            & 48 & 43 & 24 \\
        \bottomrule
    \end{tabular}
    \caption{Trajectory-level under- and over-observation rates across $100$ samples problems from \textsc{L2} and \textsc{L3}.}
    \label{tab:over-under}
\end{table}

\subsection{Disentangling Awareness from Capability}

A natural follow-up is to ask whether the under- and over-observation we identify can be removed by simply telling the agent how to observe. Concretely, we want to disentangle two hypotheses, (i) can a model's observation behavior be steered through prompting at all, and, if so, (ii) can the specific under- and over-observation problem in living-screen-native settings be resolved by prompting the model to follow the human observation strategy?

\paragraph{Setup}
We design three prompt variants on top of the default system prompt. \textit{Watch-Less} explicitly instructs the agent to keep observation cost minimal and to skip videos whenever possible, \textit{Watch-More} instructs the agent to watch videos closely, and \textit{Mimic-Human} verbalizes the human strategy distilled in Figure \ref{fig:obs-behavior}, which is to briefly glance at each video to gauge relevance and then commit to a longer watch only when the content materially supports the task. We run the three variants on Seed-1.8, which exhibits the most pronounced over-observation in Table~\ref{tab:over-under}, and report \textit{SR}, \textit{WR}, the under- and over-observation rates.

\begin{table}[t]
\setlength{\tabcolsep}{1.7mm}
    \centering
    \begin{tabular}{lcccc}
        \toprule
        \textbf{Prompt} & \textit{\textbf{SR$\uparrow$}} & \textit{\textbf{WR$\downarrow$}} & \textit{\textbf{Under$\downarrow$}} & \textit{\textbf{Over$\downarrow$}} \\
        \midrule
        Default        & 65.6 & 25.1 & 48 & 43 \\
        \midrule
        Watch-Less     & 56.3 & 19.3 & 59 & 32 \\
        Watch-More     & 57.1 & 32.2 & 40 & 69 \\
        Mimic-Human    & 59.4 & 23.6 & 50 & 36 \\
        \bottomrule
    \end{tabular}
    \caption{Effect of three prompt-level interventions on Seed-1.8.}
    \label{tab:awareness}
\end{table}

\paragraph{Prompting shifts behavior only}
Table \ref{tab:awareness} answers the two questions in turn. For (i), prompting steers behavior as intended, as \textit{Watch-Less} reduces \textit{WR} from $31.8$ to $26.0$, while \textit{Watch-More} pushes \textit{WR} up to $38.9$. For (ii), however, the answer is negative. \textit{Mimic-Human} barely moves efficiency metrics, with \textit{SR} in fact dropping. Notably, none of the three variants improves \textit{SR} over the default. This is an evidence that over- and under-observation are not awareness failures that prompting can fix, but a genuine capability deficit in current MLLMs.

\section{Conclusion}

We introduce Living-Screen-Native GUI agents  that operate on screens evolving in continuous time and decide for themselves what visual slice of the screen to observe. We then instantiate the setting on short-video platforms as \benchname{}, the first benchmark to combine a faithful browser-based environment, a three-tier task suite, and metrics that jointly score accuracy and information efficiency. None of the evaluated models reaches the human cost-accuracy performance, and their dominant failure mode is over- and under-observation, where models systematically watch either too much or too little, in model-specific but consistently miscalibrated ways. The results position observation control as a novel and currently underdeveloped axis of GUI-agent capability, alongside action grounding and content understanding. We hope \benchname{} will serve as a stepping stone toward agents that can inhabit truly living interfaces.

\section*{Limitations}

\paragraph{Platform replica}
To ensure controllability and reproducibility, \benchname{} is built on a high-fidelity browser-based replica of a short-video application rather than on a deployed platform. While the replica faithfully reproduces the on-screen affordances that drive the living-screen-native setting, it does not model a live recommendation engine, account-level personalization, or social-graph effects. Our findings therefore characterize agent behavior in the controlled replica.

\paragraph{Language and cultural coverage}
All videos, comments, and task instructions in \benchname{} are in Chinese, and the visual styling reflects a single short-video product ecosystem. Although we believe the over- and under-observation phenomenon is rooted in agent capability rather than language, its precise magnitude on other languages and other short-video platforms is not directly evaluated and may differ.

\paragraph{Annotation scale}
The final \benchname{} contains $499$ tasks over $1{,}528$ unique videos, and the human study of over- and under-observation is based on $100$ trajectories per agent. The scale is intermediate among single-video QA benchmarks, but reflects the substantially higher per-task cost of our annotation pipeline. The feed curation, task and answer writing, shortcut filtering, and cross-annotator review together amount to roughly $1$ person-hour per task that survives into the final test set. We see this person-hour-per-task number as a strong signal of data quality.

\section*{Ethical Considerations}

\paragraph{Dual use of autonomous agents.}
The living-screen-native setting is motivated by, and brings closer, the prospect of highly autonomous agents operating on real-world content platforms for content discovery, accessibility, moderation support, etc. Such agents carry clear social benefits but also non-trivial risks, including misuse for automated misinformation amplification, evasion of moderation, or fake-engagement generation. \benchname{} is intended as an evaluation tool that helps the community measure and stress-test such capabilities before deployment, and we recommend that any downstream system trained or evaluated on it be paired with human-in-the-loop oversight prior to real-platform use.

\paragraph{Data licensing and desensitisation.}
All videos and associated metadata in \benchname{} are sourced from publicly released, openly licensed short-video datasets, used within their original licence terms. We therefore do not introduce additional data-sourcing risks beyond those already cleared by the underlying public datasets.

\paragraph{Annotator considerations.}
All annotators in this work are co-authors of the paper and contributed annotation as part of their research participation. Because a subset of tasks involves content-moderation scenarios that may surface distressing material (e.g., violence, low-quality content), annotators were briefed on the content scope in advance, were free to skip any item they were uncomfortable with, and worked in self-paced sessions.

\bibliography{custom}

\appendix

\appendix

\section{Prompts}
\label{sec:appendix-prompts}

Because all videos, comments, and task instructions in \benchname{} are in Chinese, every prompt actually fed to the models is also written in Chinese. For readability, we present the English translation of each prompt first, followed by the original Chinese version.

\subsection{Main Agent System Prompt}
\label{sec:appendix-prompts-main}

This prompt is prepended to every agent rollout in the main experiments and the agent-design ablations.

\paragraph{English translation.}
\begin{quote}\itshape
You are a professional GUI control agent specialised in interacting with a short-video platform. Your task is to follow a long-horizon user goal and output precise instructions based on observed screenshots or video streams.

Your input images come from screenshots of a typical short-video feed interface. The feed is of finite length; after several consecutive upward swipes the interface may stop updating. Use \textbf{relative} image coordinates: treat the width and height of the input image as a normalised coordinate system from 0 to 1000, so that the top-left corner is \texttt{<point>0 0</point>}, the top-right is \texttt{<point>1000 0</point>}, and the bottom-right is \texttt{<point>1000 1000</point>}. You may click on the video to pause/resume, click on a specific position of the progress bar to seek, swipe upward to advance to the next video, or swipe downward to return to the previous video.

After your reasoning in each response, you should call a tool to act, and call the \texttt{finish} tool when the task is fully completed.
\end{quote}

\paragraph{Original Chinese.}
\begin{quote}\itshape
\begin{zh}
你是一个专业的 GUI 操控智能体,专门负责在短视频平台中进行交互。你的任务是根据用户的长期目标,通过观察屏幕截图或视频流输出精确指令。

你的输入图像来自于一个典型的短视频流环境界面的截图。该短视频流的长度是有限的,如果连续多次向上滑动后界面可能不再更新内容。请使用\textbf{相对}图像坐标,将输入图像的宽高视为 0 到 1000 的归一化坐标系:图像左上角为 \texttt{<point>0 0</point>},右上角为 \texttt{<point>1000 0</point>},右下角为 \texttt{<point>1000 1000</point>}。你可以进行点击视频以暂停/继续、点击进度条特定位置跳转、从下向上拖拽进入下个视频和从上向下拖拽回到上个视频等操作。

每次响应中的推理思考后,你应调用工具进行操作,并在任务圆满完成时调用 \texttt{finish} 工具。
\end{zh}
\end{quote}

\subsection{Observation-Behaviour Intervention Prompts}
\label{sec:appendix-prompts-obs}

The following three instructions are appended to the main system prompt in the analysis section to investigate observation behavior control. All other components remain unchanged.

\paragraph{Watch-Less --- English translation.}
\begin{quote}\itshape
Be economical with observation. Avoid invoking \texttt{watch} unless strictly necessary, and prefer to \texttt{wait} or skip past a video. When in doubt about whether to keep watching, choose to move on rather than continue observing.
\end{quote}

\paragraph{Watch-Less --- Original Chinese.}
\begin{quote}\itshape
\begin{zh}
请节约你的观察开销。除非确有必要,不要调用 \texttt{watch},优先使用 \texttt{wait} 或直接跳过视频。当你犹豫是否要继续看时,选择前进而非继续观察。
\end{zh}
\end{quote}

\paragraph{Watch-More --- English translation.}
\begin{quote}\itshape
Observe thoroughly. Whenever a video appears potentially relevant to the task, invoke \texttt{watch} to inspect it closely, and prefer to watch longer segments over shorter ones. Do not rush to answer until you have closely examined the videos relevant to the answer.
\end{quote}

\paragraph{Watch-More --- Original Chinese.}
\begin{quote}\itshape
\begin{zh}
请充分地进行观察。只要一个视频与任务可能相关,就调用 \texttt{watch} 仔细查看,且倾向于观察更长的片段而非更短的片段。在详细查看过与答案相关的视频之前,不要急于作答。
\end{zh}
\end{quote}

\paragraph{Mimic-Human --- English translation.}
\begin{quote}\itshape
Observe in the way an expert, attentive human user would. For each video, first take a brief look of a few seconds to judge whether it is relevant to the task; only when this initial look suggests that the video materially supports your answer should you invoke \texttt{watch} for a short additional observation; if you suspect there is further relevant content not yet covered, then try watching a later part of the video.
\end{quote}

\paragraph{Mimic-Human --- Original Chinese.}
\begin{quote}\itshape
\begin{zh}
请模仿专业、专注的人类用户的方式进行观察。对每一个视频,先简短查看几秒,以判断它是否与任务相关;只有当这一次初步查看表明该视频对回答任务有实质帮助时,才进一步使用 \texttt{watch} 短时间查看;如果发现可能有更多未掌握的相关内容,再尝试观看更靠后的部分。
\end{zh}
\end{quote}

\subsection{Shortcut-Filter Prompt}
\label{sec:appendix-prompts-filter}

This prompt is fed to a text-only LLM during quality control to filter out tasks that can already be solved from non-video metadata alone. Placeholders \texttt{\{meta\_text\}} and \texttt{\{question\}} are filled with the concatenated metadata of the candidate feed and the authored question respectively.

\paragraph{English translation.}
\begin{quote}\itshape
The following is the metadata of some videos:

\texttt{\{meta\_text\}}

Please answer the following question based solely on the metadata above:

Question: \texttt{\{question\}}

\textbf{Requirements:}
\begin{itemize}\itemsep0pt
  \item Output only the letter of the answer (e.g., A, B, C, D).
  \item Do not output any other text.
  \item If the question cannot be answered, output ``Cannot answer''.
\end{itemize}
\end{quote}

\paragraph{Original Chinese.}
\begin{quote}\itshape
\begin{zh}
以下是一些视频的元信息:

\texttt{\{meta\_text\}}

请仅根据上述元信息回答以下问题:

问题:\texttt{\{question\}}

\textbf{要求:}
\begin{itemize}\itemsep0pt
  \item 只输出答案字母（如 A、B、C、D 等）
  \item 不要输出任何其他文字
  \item 如果无法回答,输出 ``无法回答''
\end{itemize}
\end{zh}
\end{quote}

\section{Benchmark Details}
\label{sec:appendix-benchmark}

\subsection{Platform Overview}
\label{sec:appendix-platform-overview}

\begin{figure}[htbp]
    \centering
    \includegraphics[width=1.0\linewidth]{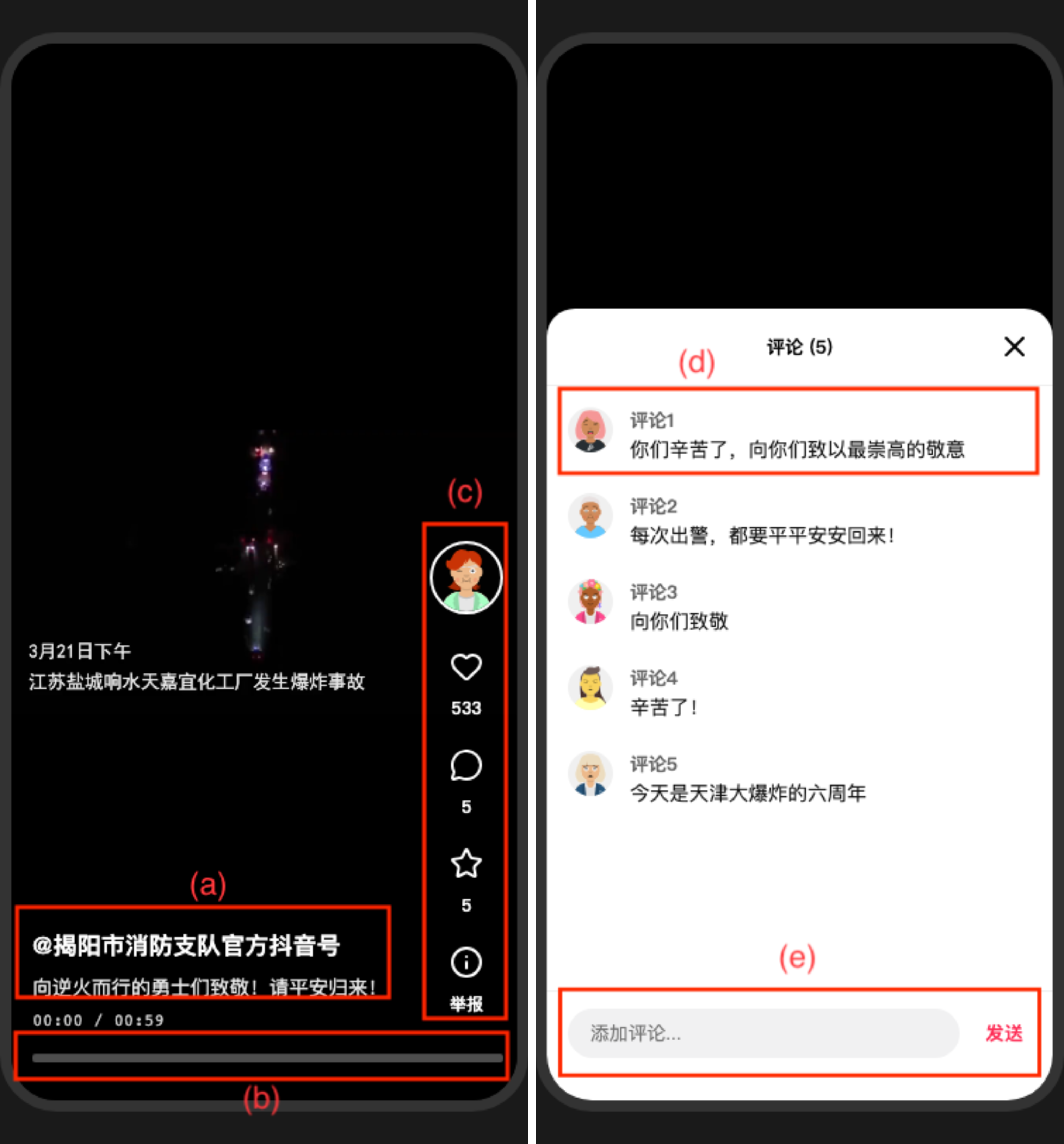}
    \caption{Overview of the \benchname{} platform with annotated UI components. (a) author meta panel, (b) progress bar, (c) side action rail (like / comment / collect / report), (d) comment drawer, (e) comment box.
    }
    \label{fig:appendix-ui}
\end{figure}

Figure~\ref{fig:appendix-ui} shows the full \benchname{} interface together with annotations of the major UI components agents can interact with. The viewport is fixed at a typical mobile resolution (% TODO: WxH px
\,XX$\times$XX\,px) and reproduces the layout of a modern short-video application, including the autoplaying video region, the progress bar, the side action rail (like, comment, collect, share), the comment drawer, and the swipe-to-next-video gesture region. All components are rendered with real CSS/JS rather than emulated, so that agent screenshots and recordings match what a human user would see in the same interface.

\subsection{Platform Implementation}
\label{sec:appendix-platform-impl}

The \benchname{} platform is implemented as a self-hosted web application. The front end is built with HTML, CSS, and vanilla JavaScript, rendering each task's feed as a vertically scrollable stack of video cards with the same affordances as a mobile short-video client. The back end is a lightweight \textsc{Flask} server that serves the page, video assets, and metadata, and maintains per-episode environment state (e.g., which videos have been liked, collected, or reported) so that L1 and L3 tasks can be auto-graded against a declarative ground-truth specification.

Agent--platform interaction is mediated by \textsc{Playwright}, which controls a headless Chromium instance and exposes the action API described in Appendix~\ref{sec:appendix-action-api}. Every primitive in the action space is translated by the wrapper into the underlying Playwright call (mouse click at pixel $(x,y)$, scroll wheel, keyboard input, etc.), so that the agent's view of the platform is restricted to rendered screenshots and recorded video clips and never includes DOM, accessibility tree, or raw video file access.

\section{Agent and Evaluation Details}
\label{sec:appendix-agent}

\subsection{Action API Specification}
\label{sec:appendix-action-api}

We list the full JSON schema of every primitive in the action space in Listing \ref{lst:tool-schema}.

\begin{lstlisting}[
    language=json,
    caption={JSON schema of agent action space},
    label={lst:tool-schema},
    float=false
]
[
  {
    "name": "mark_point",
    "description": "Place a marker on the screen for later reference.",
    "parameters": {
      "x":     {"type": "integer", "range": [0, 1000], "required": true},
      "y":     {"type": "integer", "range": [0, 1000], "required": true},
      "label": {"type": "string",  "required": true}
    }
  },
  {
    "name": "click",
    "description": "Click on the screen at the given normalised coordinate.",
    "parameters": {
      "x": {"type": "integer", "range": [0, 1000], "required": true},
      "y": {"type": "integer", "range": [0, 1000], "required": true}
    }
  },
  {
    "name": "swipe",
    "description": "Swipe from a start point to an end point on the screen.",
    "parameters": {
      "x1": {"type": "integer", "range": [0, 1000], "required": true},
      "y1": {"type": "integer", "range": [0, 1000], "required": true},
      "x2": {"type": "integer", "range": [0, 1000], "required": true},
      "y2": {"type": "integer", "range": [0, 1000], "required": true}
    }
  },
  {
    "name": "type",
    "description": "Type the given text into the currently focused input field.",
    "parameters": {
      "text": {"type": "string", "required": true}
    }
  },
  {
    "name": "press",
    "description": "Press a single keyboard key.",
    "parameters": {
        "type": "object",
        "properties": {
            "key": { "type": "string", "required": true}
        },
  },
  {
    "name": "wait",
    "description": "Pause without consuming the video stream; the screen continues to update autonomously.",
    "parameters": {
      "seconds": {"type": "number", "range": [0, 60], "required": true}
    }
  },
  {
    "name": "watch",
    "description": "Actively observe the video stream for a fixed duration at the specified sampling rate.",
    "parameters": {
      "seconds": {"type": "number", "range": [0, 60],     "required": true},
      "fps":     {"type": "number", "range": [0.1, 30],   "default": 1}
    }
  },
  {
    "name": "answer",
    "description": "Submit the final answer for understanding-style (L2) tasks and terminate the episode.",
    "parameters": {
      "content": {"type": "string", "required": true}
    }
  },
  {
    "name": "finish",
    "description": "Signal that the task has been completed and terminate the episode.",
    "parameters": {}
  }
]
\end{lstlisting}

% \begin{table}[htb!]
%     \centering
%     \begin{tabular}{c >{\raggedright\arraybackslash}p{0.7\linewidth}}
%         \toprule
%         \textbf{Function} & \textbf{Description} \\
%         \midrule
%         \textsc{mark}        & Mark a normalized $(x,y)$ pixel coordinate. \\
%         \textsc{click}       & Click at a $(x,y)$ coordinate. \\
%         \textsc{swipe}       & Swipe from $(x_1,y_1)$ to $(x_2,y_2)$. \\
%         \textsc{type}        & Type a string into the currently focused input box. \\
%         \textsc{press}       & Press a single keyboard key. \\
%         \textsc{watch}       & Record the on-screen rendering for $\Delta t$ seconds. \\
%         \textsc{wait}        & Let the screen evolve for $\Delta t$ seconds without watching. \\
%         \textsc{answer}      & Emit the final textual answer and terminate the episode. \\
%         \bottomrule
%     \end{tabular}
%     \caption{Action space of our agent design.}
%     \label{tab:function}
% \end{table}

\subsection{Video Input Format}
\label{sec:appendix-video-input}

The format of visual feedback returned to the agent depends on which primitive is executed.

\paragraph{Non-\texttt{watch} actions.} For all GUI primitives (\texttt{click}, \texttt{swipe}, \texttt{type}, \texttt{mark\_point}) and the temporal \texttt{wait}, the environment returns a single screenshot of the page captured immediately after the action settles. This is the standard single-frame observation used by virtually all prior GUI agents.

\paragraph{\texttt{watch} action.} The \texttt{watch} primitive instead records the screen for the requested duration at the specified frame rate. The resulting clip is delivered to the model in one of two ways depending on its input modality. If the backbone natively ingests video, the recorded clip is encoded as base64 and passed directly through the video input channel, preserving the original frame rate and temporal continuity. For backbones that accept only images, we uniformly subsample the clip at the same frame rate and feed the resulting frames as a multi-image input in temporal order.

\end{document}